\documentclass[10pt,twocolumn,letterpaper]{article}
\usepackage{tikz}
\usepackage{pgfplots}
\usepackage{verbatim}
\usepackage{xcolor}
\usepackage{cvpr}
\usepackage{times}
\usepackage{epsfig}
\usepackage{graphicx}
\usepackage{amsmath}
\usepackage{amssymb}
\usepackage{float}
\usepackage[linesnumbered,ruled]{algorithm2e}
\usepackage{graphicx}
\usepackage{amsmath,amssymb} % define this before the line numbering.
\usepackage{color}
\usepackage{tikz}
\usepackage{pgfplots}
\pgfplotsset{compat=newest}
\usepackage{verbatim}
\usepackage{xcolor}
\usepackage{pdfpages} 
\usepackage{times}
\usepackage{epsfig}
\usepackage{float}
\usepackage{tikz}
\usepackage{pgfplots}
\usepackage{verbatim}
\usepackage{xcolor}
\usepackage{times}
\usepackage{epsfig}
\usepackage{graphicx}
\usepackage{amsmath}
\usepackage{amssymb}
\usepackage{float}
\usepackage[linesnumbered,ruled]{algorithm2e}
\usepackage{subfigure}

\cvprfinalcopy 
\def\cvprPaperID{1542}

\begin{document}

\title{FAME:  \\
Face Association through Model Evolution}

\author{Eren Golge\\
Bilkent University\\
06800 Ankara/Turkey\\
{\tt\small eren.golge@bilkent.edu.tr}
\and
Pinar Duygulu\\
Bilkent University\\
06800 Ankara/Turkey\\
{\tt\small pinar.duygulu@gmail.com}
}
\maketitle

\begin{abstract}

We attack the problem of learning face models for public faces from weakly-labelled images collected from web through querying a name. The data is very noisy even after face detection, with several irrelevant faces corresponding to other people. We propose a novel method,  {\bf Face Association through Model Evolution (FAME)}, that is able to prune the data in an iterative way, for the face models associated to a name to evolve.  
The idea is based on capturing discriminativeness and representativeness of each instance and eliminating the outliers. The final models are used to classify faces on novel datasets with possibly different characteristics.  On benchmark datasets, our results are comparable to or better than state-of-the-art studies for the task of face identification. 

%Experimental results on benchmark datasets are competitive to state-of-the-art studies.  
 
 % and it does not require any human hand other than the provided query to the system.
 % In this setting, we detect single, high confident face from each returned image and curate our raw dataset. 
 %However this data is akin to include irrelevant or outlier detections as well.
%  We propose a novel algorithm that is able to prune the data in a iterative formation up to a desired level by using a large set of negative instance. 
%The idea is to capture determinativeness and the representativeness of each instance in the given data. First we learn a linear model that separates the data from the large negative set. The confidence of each instance is assumed to be the score of discriminativeness. We select half number of instances with higher confidence values as the control set and label the remaining instances are candidates to eliminate. We learn another linear model between these two sub-sets. This model is assumed to be the measure of representativeness of the data. Then we eliminate $m$ highly confident negatives of the last model as possible outliers. This iteration goes on up until a desired level and we supposedly end up with a pure data for each name to learn final models of classification. Those models are tested on some benchmark datasets as a testimony of our method and observed comparative results to the state-of-art.
\end{abstract}

%------------------------------------------------------------------------- 
\section{Introduction}

%, such as  pose, illumination, expression, make-up, hair-style, clothing and age variations. 
To label faces of friends in social networks or celebrities and politicians in news, automatic methods are indispensable to manage large number of face images piling up on the web.
On the other hand, unlike their counterparts in controlled datasets, faces on the web inherit all type of challenges naturally, resulting in the traditional methods incapable to recognise.

Recent availability of real-world face datasets \cite{Berg-CVPR2004,Kumar-2009} accelerated the works in web-scale face verification, that is  given a pair of faces deciding their identity. On the other hand, identification, that is finding the identity of a face, is still relatively less studied for the real-world faces. The requirement for a considerable amount of faces to be labeled is the main bottleneck for scalability in identification. Continuous inclusion of new individuals, and new instances for each individual should also be considered for web-scale identification task.

In this study, we challenge the identification of faces for famous people. The famous people tend to change their make-up, hair style/colour, and accessories more often compared to regular people, resulting in large number of varieties in face images.   Moreover, they are likely to appear with others in photographs, causing faces of irrelevant people to be retrieved.

We propose a new method, {\bf FAME}, that utilises the noisy results obtained through a name query to construct models in identifying famous people.  Our models evolve through consecutive iterations to associate the query name with the correct set of faces. These models are then used to label faces on novel datasets.  
FAME removes the outlier faces in constructing models,  while retaining the diversity as much as possible.
Details of FAME will follow the review of recent work on relevant domains.
 
\section{Related Work}
{\bf Naming faces using weakly-labeled data: } The work of Berg \etal is one of the first attempts in labelling large number of faces from weakly-labeled web images \cite{Berg-CVPR2004, Berg-NIPS2004} with the "Labeled Faces in the Wild" (LFW) dataset introduced.  It is assumed that in an image at most one face can correspond to a name, and names are used as constraints in clustering faces. Appearances of faces are modelled through Gaussian mixture model with one mixture per name. In \cite{Berg-CVPR2004}, k-PCA is used to reduce the dimensionality of the data and LDA is used for projection. Initial discriminant space learned from faces with a single associated name is used for clustering through a modified k-means. Better discriminants are then learned to re-cluster. In \cite{Berg-NIPS2004} face name associations are captured through an EM based approach.
% and it is also shown that language model helps to improve appearance only based models.

For aligning names and faces in an (a)symmetric way, Pham \etal \cite{Pham-TransMM2010} cluster the faces using a hierarchical agglomerative clustering method. They use the constraint that faces in an  image cannot be in the same cluster. They then use an EM based approach for aligning names and faces based on probability of reoccurrences. They use a 3D morphable model for face representation.
They introduce the picturedness and namedness: the probability of a person being in the picture based on textual info, and being in the text based on visual info. 

Ideally, there should be a single cluster per person. However, these methods are likely to produce clusters with several people mixed in, and multiple clusters for the same person. 

In  \cite{Ozkan-CVPR2006, Ozkan-PR2010}, Ozkan and Duygulu consider the problem as retrieving faces for a single query name, and then pruning the set from the irrelevant faces.  A similarity graph is constructed where the nodes are faces, and edges are the similarity between faces. With the assumption that the most similar subset of faces will correspond to the queried name, the densest component in the graph is sought using a greedy method.  In \cite{Guillaumin-CVPR2008}, the method of  \cite{Ozkan-CVPR2006, Ozkan-PR2010} is improved by introducing the constraint for each image to contain a single instance of the queried person and replacing the threshold in constructing the binary graphs with assigning non-zero weights to k nearest neighbours. The authors further generalised the graph based method for for multi-person naming, as well as null assignments. They propose a min-cost max-flow based approach to optimise face name assignments under unique matching constraints. 
%In \cite{Guillaumin-ICCV2009}. a logistic discriminant approach which learns the metric from pairs of faces  is proposed for identification of faces. As another approach for face identification, they propose a method  where the probability of two faces belonging to the same class is computed in a nearest neighbour based approach.

%Guillaumin \etal proposed a set of other methods for face naming with caption based supervision.
In \cite{Guillaumin-ECCV2010} face-name association problem is tackled as a multiple instance learning problem over pairs of bags. Detected faces in an image is put into a bag, and names detected in the caption are put into the corresponding set of labels. A pair of bags is labeled as positive if they share at least one label, and negative otherwise. The results are reported on Labelled Yahoo! News dataset which is obtained through manually annotating and extending LFW dataset. In \cite{Guillaumin-IJCV2012}, it is shown that the performance of graph-based and generative approaches for text-based face retrieval and face-name association tasks can be improved with the incorporation of logistic discriminant based metric learning (LDML) \cite{Guillaumin-ICCV2009}.

Kumar \etal \cite{Kumar-2009} introduced attribute and smile classifiers for verifying the identity of faces. For describable aspects of visual appearance, binary attribute classifiers are trained with the help of AMT. 
Moreover, simile classifiers are trained to recognise the similarity of faces to
specific reference people. Pub-Fig, dataset of public figures on the web, is presented alternative to LFW with larger number of individuals each having more instances.

% Pham et al. \cite{Pham-MM2011} use the idea of label propogation, to name unlabelled faces in videos starting from a set of seed labeled faces. Together with visual similarities, they also make use of constraints for assigning a single name to face tracks and not labelling two faces in a single frame with the same name.

%In  \cite{Guillaumin-IJCV2012}, the concept of "friends" is introduced for query expansion. The names of the people frequently co-occurring with the queried person is used for extending the set of faces, and resemblance of the faces to the friends is used for better modelling of the query person.   

Recently, PubFig83, a subset of PubFig dataset with near-duplicates eliminated and individuals with large number of instances are selected,  is provided for face identification task \cite{pinto2011scaling}.   
Inspired from biological systems,  Pinto {\em et al.} \cite{pinto2011scaling}  consider V1-like features and introduce both single- and multi-layer feature extraction architecture followed by LinearSVM classifier.% In \cite{chiachia2012person},  person specific partial least squares (PS-PLS) approach is presented to generate subspaces for familiar faces, such as celebrities.

\cite{Ortiz-CVIU2014} define the open-universe face identification problem as identifying faces with one of the labeled categories in a dataset including distractor faces that do not belong to any of the labels.
In \cite{Becker-CVPRW2013}, the authors combine PubFig83, as being the set of labeled individuals,  and LFW, as being the set of distractors. On this set, they evaluate a set of identification methods including nearest neighbour, SVM, sparse representation based classification (SRC) and its variants, as well as linearly approximated SRC that they proposed in \cite{Ortiz-CVIU2014}.

Other recent work include  \cite{Simonyan-BMVC2013} where Fisher vectors on densely sampled SIFT features are utilised. Large margin dimensionality reduction is used to reduce high dimensionality.

{\bf Harvesting web for concept learning:}
Recently, there have been many studies on harvesting web for re-ranking of search results and building qualified training sets \cite{Fergus-ICCV2005, Berg-CVPR2006, Berg-CVPRW2009, Fan-CVPR2010, Li-IJCV2010, Schroff-PAMI2011, Chen-ICCV2013}. 
In \cite{Berg-CVPR2006} visual features and surrounding the text are used for collecting animal images from web, and visual exemplars are obtained through clustering text. Relevant clusters are required to be  identified manually, as well as irrelevant images in clusters. 
In  \cite{Li-IJCV2010},  OPTIMOL framework is presented to incrementally learn object categories from web search results. Given a set of seed images a non parametric latent topic model is applied to categorise collected web images. The model is iteratively updated with the newly categorised images. To prevent over specialised results, a set of cache images with high diversity are retained at each iteration. 
In  \cite{Schroff-PAMI2011} after the removal of abstract images from the search results collected through text and image search, text and metadata are used to re-rank the images. A visual classifier is trained by sampling from the top ranked images as positives and random images from other categories as negatives. 
Recently,  NEIL  \cite{Chen-ICCV2013} is proposed  to learn object and scene categories, as well as common sense knowledge using web search results.
{\bf Discovering representative and discriminative instances:}
Our method is also related to the recently emerged studies in discovering discriminative patches. 
~\cite{liharvesting, kim2009unsupervised, Singh-ECCV2012, doersch2012makes, doersch2013mid, jain2013representing, endres2013learning, juneja2013blocks}. 
In \cite{Singh-ECCV2012}, discriminative patches in images are discovered through an iterative method 
which alternates between clustering and training discriminative classifiers.
Li \etal \cite{liharvesting} solves same problem with multiple instance learning. \cite{doersch2012makes} and \cite{juneja2013blocks} apply the idea to scene images for learning discriminative properties by embracing the unsupervised exemplar models. Moreover \cite{doersch2013mid} enhances the unsupervised learning schema by more robust alternation of Mean-Shift clusteringalgorithm. Disciminative patch ideas is also applied to video domain by \cite{jain2013representing}. 
\section{Our approach}
\begin{figure}
\centering
\includegraphics[width=.45\textwidth]{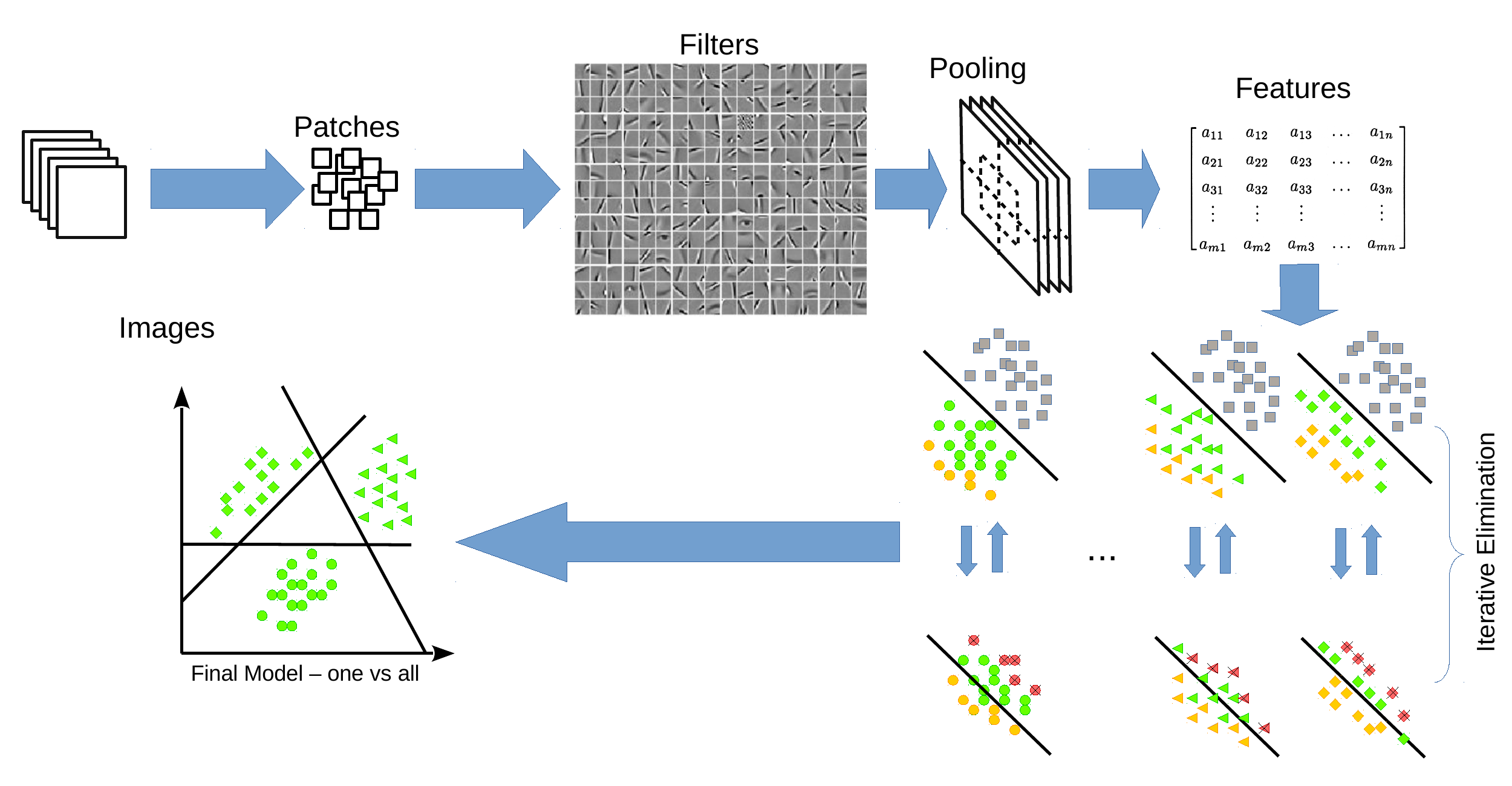}
\caption{Overview of the proposed method.}
\label{fig:overview}
\end{figure}

An important caveat in learning models from weakly-labelled data is the impurity of the collection. To be useful, spurious instances should be eliminated before generating models for each category.  In this study, we present a new approach for learning better models through iteratively pruning the data (see Figure\ref{fig:overview}).  
First, we benefit from large number of global negatives representing the rest of the world against the class of interest. Next, among the candidate in-class examples we try to separate the most confident instances from the others. These two successive steps are repeated to eliminate outlier instances iteratively.
To consider intra-class variability, we use a representation that results in large dimensional feature vectors to make each class linearly separable even when the data include some level of variation. 
The model evolution and representation will be detailed in the following.
\subsection{Model Evolution}
We propose a method that allows the models to evolve through eliminating the outlier instances with successive linear classifiers. First, we learn a hyperplane that separates the initial set of candidate class instances from the large set of global negatives. Global negative set is curated by the instances of other classes and the random face images collected from Web. 
Then, we select some fraction of the class instances that are distant from the separating hyperplane. We use these instances as the discriminative seed set, since they are confidently classified against the rest of the world.  We consider the rest of the class data as possible negatives. We then learn another model that try to capture in-class dissimilarities between discriminative examples and possible negatives. At the final step, we combine the confidence scores of the first and the second models. By combining the two scores,  that respectively correspond to the confidence of being different from the rest of the world, and in-class affinity of the instance,  we get a measure of instance saliency. Over these confidence scores we detect instances with the lowest scores as the outliers for that iteration. These steps are iterated multiple times up to a desired level of pruning. The representation that we use (see Section\ref{sec:representation}) might cause computational burden with complicated learning models. Therefore, we leverage simple linear regression (LR) models with L1 norm regularisation performing sparse feature selection as the learning evolves.
Sparsity makes categories more distinct and captures category related commonalities. 

Algorithm \ref{algo:FAME} summarises  our data elimination procedure.
$C=\{c_1,c_2, \ldots c_m\}$ refers to the examples collected for a class and  $N=\{n_1,n_2,...,n_l\}$ refers to the the vast numbers of global negatives. Each vector is a $d$ dimensional representation of a single face image. At each iteration $t$, the first LR model $M^1$  learns a hyperplane between the candidate set of class instances $C$ and global negatives $N$. 
Then the current $C$ is divided into two subsets: $p$ instances in $C$ that are farthest from the hyperplane are kept as the candidate positive set ($C^+$) and the rest is considered as the negative set ($C^-$) for the next model. $C^+$ is the set of salient instances for the class and $C^-$ is the set of possible spurious instances. The second LR model $M^2$ uses $C^+$ as positive and $C^-$ as the negative set to learn best possible hyperplane separating them. For each instance in $C^-$, by aggregating the confidence values of both models, we eliminate $o$ instances with the lowest scores as the outliers. At the next iterations, we run all the steps again and end up with a clean set of class instances $C$.

This iterative procedure continues until it satisfies a stopping condition which is refined by $M^1$'s training accuracy as the measure of present data quality. As we incrementally remove poor instances, we expect to have better separation against the negative instances therefore $M^1$'s accuracy increases. However, if the accuracy saturates or degrade then we stop the algorithm.
Alternatively, when we have very large number of class instances, we can divide data into two independent subset and apply the iterative elimination to both as we measure the quality of one set's $M^1$ over  the other set's $C$ at each iteration $t$. It is similar to co-training approach and more robust to over-fitting, albeit it requires very large number of instances for convincing results. 
\begin{algorithm}
\begin{small}
\DontPrintSemicolon % Some LaTeX compilers require you to use \dontprintsemicolon instead 
In the real code we use vectorized implementation whereas we write down iterative pseudo-code for the favour of simplicity.\\
\KwIn{$C$, $N$, $o$, $p$} 
\KwOut{$C$} 
%$C^+ \gets \emptyset$\;
%$C^- \gets \emptyset$\;
$C_0 \gets C$\;
$t \gets 1$\;
\While{$stoppingConditionNotSatisfied()$}{
  %$\sigma^t \gets computeWindow(t, \sigma^{init}$)\;
  $M_{t}^1 \gets LogisticRegression(C_{t-1} , N$)\;
  $C_t^+ \gets selectTopPositives(C_{t-1}, M_{t}^1, p)$\;
  $C_t^- \gets C_{t-1} - C_t^+$\;
  $M_{t}^2 \gets LogisticRegresstion(C_t^+,C_t^-)$\;
  $[ S_1^{-}, S_2^{-} ] \gets getConfidenceScores(C_t^-, M_{t}^1 M_{t}^2)$\;
  $O_t \gets selectOutliers(C_t^-, S_1^{-}, S_2^{-}, o)$\;
  $C_t \gets C_{t-1} - O_t$\;
  $t \gets t+1$\;
}
$C \gets C_t$\;
\Return{$C$}\;
\caption{FAME} 
\label{algo:FAME}
\end{small}
\end{algorithm}
\subsection{Representation}
\label{sec:representation}
\begin{figure}
\begin{center}
	\subfigure[]{\label{fig:cent_filters} \includegraphics[scale=0.2]{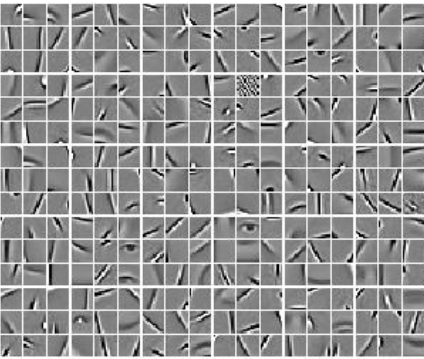}}
	\subfigure[]{\label{fig:lbp_filters} \includegraphics[scale=0.18]{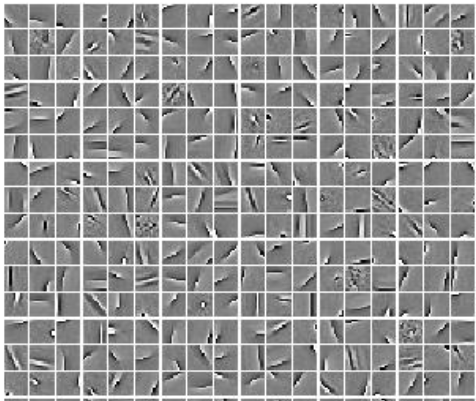}}	
	\subfigure[]{\label{fig:outlier_filters} \includegraphics[scale=0.35]{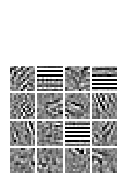}}
	\subfigure[]{\label{fig:lbp_images} \includegraphics[scale=0.35]{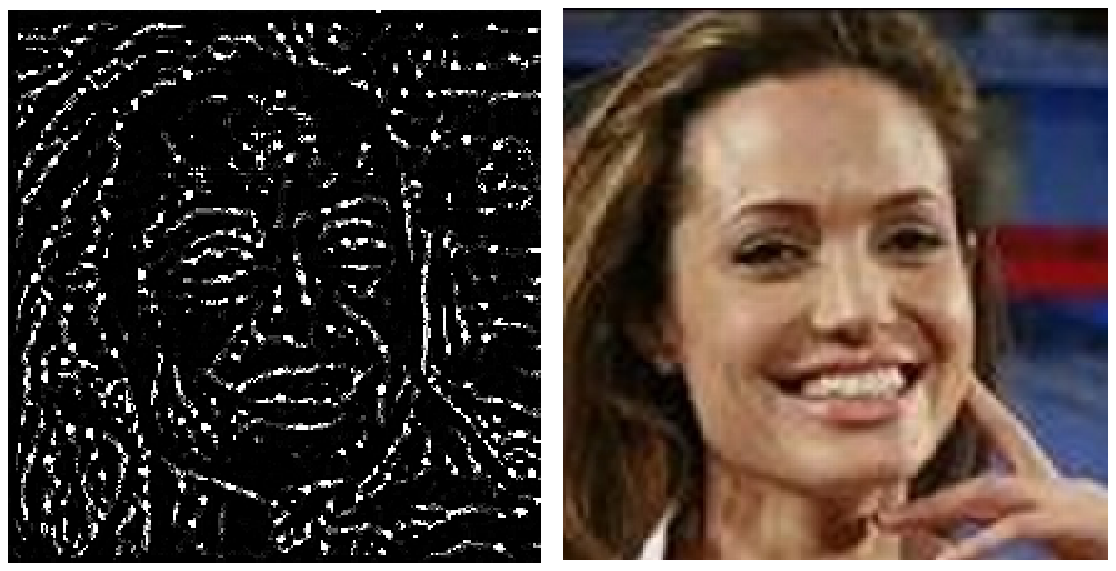} }	
	\caption{Random set of filters learned from (a) whitened raw image pixels, and (b) LBP encoded images. (c) Outlier filters of raw-image filters. (d) LBP encoding of a given RGB image. We might observe eye or mount shaped filters from the raw image filters and more textural information from the LBP encoded filters. Outlier filters are very cluttered and observe low number of activations mostly from background patches.}
\end{center}
\end{figure}

To represent face images we learn two distinct set of filters by an unsupervised method similar to \cite{coates2011analysis} (see Figure\ref{fig:outlier_filters} ). First set is learned from the raw-pixel random patches extracted from grey-scale images. The second set is learned from LBP \cite{ahonen2006face} encoded images. First set of learned filters are receptive to edge- and corner-like structural points and the second set is more sensitive to textural commonalities of the LBP histogram statistics.LBP encoded images are invariant to illumination since  the intensity relations between pixels are considered instead of exact pixel values. 
We use rotation invariant LBP encoding \cite{ojala2000gray} that gives binary codes for each pixel. Then, we convert these binary codes into corresponding integer values. A Gaussian filter is used to smooth out the heavy-tailed locations. 
The pipeline in order to learn filters from both raw-pixel and LBP images is as follows. First we extract a set of randomly sampled patches in the size of predefined receptive field. Then contrast normalisation is applied to each patch (for only raw-image filters) and  patches are whitened to reduce the correlations among dimensions. These patches are clustered using k-means into $K$ groups. We perform thresholding to centroids with box-plot statistics over the activations counts to remove the outlier centroids that are supposedly not representative for the face images but background clutters. After the learning phase, centroid activations are collected from receptive fields with small striding. We applied spatial average pooling onto five different grids including a grid at the center of the image additional to 4 equal-sized quadrants since face images includes important spatial regularities at the center. We use triangular activation function to map each receptive field to learned centroids. This yields a $5xK$ dimensional representation for each face. However, since we use two different set of filters, at the end, each image presented by $2x5xK$ dimensions.  
Thresholding of centroid activations provides a implicit removal of outlier patches as well as the salient set of centroids. We use those outlier centroids to eliminate patches at the feature extraction step by assuming the patches assigned to outlier centroids are not relevant thus avoiding them in pooling.
\section{Experiments}
\subsection{Datasets}
Images are collected using Bing to train models.  Then, two recent benchmark datasets, FAN-large \cite{ozcan2011large} and PubFig83\cite{pinto2011scaling}, are used for testing.{\bf Bing collection:}For a given name, 500 images are gathered using Bing image search \footnote{$https://www.bing.com/$}. Categories are chosen as the people having more than 50 annotated face images in FAN-large or PubFig83 datasets. In total, 226691 images are collected corresponding to 365 name categories in FAN-large, and 83 name categories in PubFig83. Additional 2500 face images for queries "female face", "male face", "face images" are collected to construct the global negatives. Face detector of \cite{zhu2012face} is used for detecting faces. Only the most confident detection is selected from each image to be put into the initial pool of faces associated with the name (on the average 450 faces per category). Other detections are added to global negatives.{\bf Test collection:}We use two sets from FAN-large face dataset \cite{ozcan2011large}: EASY and ALL. EASY subset includes faces larger than 60x70 pixels. ALL includes all names without any size constraint.  We use 138 names from EASY,  and 365 from ALL subsets, with  23952 and 199295 images respectively. On the average  there are 541 images for each name. We also use PubFig83\cite{pinto2011scaling} dataset, which is the subset of well-known PugFig dataset with 83 different celebrities having at least 100 images. PubFig83 is more convenient set for face identification problem with near-duplicate images and the ones that are no longer available at Internet are removed\cite{becker2013evaluating}. We shaped a controlled test environment by using PubFig83+LFW  \cite{becker2013evaluating}: extending PubFig83 with some distract images from LFW \cite{huang2007labeled} not belonging to any of the selected categories (distractors are six percent of correct instances).We use these distract images to extend our global negatives. For the controlled experiment,  we select name categories with more than 270 images and mixed them with random set of distract images.Then we apply full stack of FAME with 5-fold cross-validation. 
\subsection{Implementation Details}
The dataset is expanded with horizontally flipped images.
Before learning filters from raw-pixel images, each grey-level face image is resized to 60 pixels height and LBP images resized to 120 pixels height. LBP encoding has been done by 16 different filter orientation and at radius 2. We sample random patches from images and apply contrast normalization to only raw-pixel patches. Then, we perform ZCA whitening transform and set $\epsilon_{ZCA}$ to 0.5. We use receptive field of 6x6 regions with 1 stride and learn 2400 centroids for both raw-pixel images and LBP encoded images. Hence, we conclude 2 (raw-pixel + LBP) x 5 (pooling grids) x 2400 (centroids) dimensional feature representation of each image. For instance to centroid distances we used Euclidean Distance. We detect the outliers by a threshold at the 99\% upper whisker of the centroid activations. Our implementation of feature learning framework aggregated upon the code furnished by \cite{coates2011analysis}. For iterative elimination, we train L1 norm Logistic Regression model with \textit{Gauss-Seidel} algorithm \cite{shevade2003simple} and final classification is done with Linear SVM through \textit{grafting} algorithm \cite{perkins2003grafting} that learns sparse set of important features incrementally by using gradient information. At each FAME iteration we eliminate five images. We stop when there is no improvement on the first model accuracy. If the classifier saturates so quickly, iteration continues until 10\% of the instances are pruned.
If we encounter memory constraints due to large number of global negatives, at each iteration we sample a different set of negative instances, to provide slightly different linear boundaries that are able to detect different spurious instances.
\subsection{Evaluations}
We conduct controlled experiments over PubFig83+LFW. We select classes with at least 270 instances and inject 10\% (27 instances) noise instances. There are six classes conforming that criterion. Noisy images are randomly chosen from global negatives consisting of  "distract" set of PubFig83+LFW and FAN-large faces that we collected. As a result, we have 297x6 training instances. We apply FAME to this data while applying cross-validation at each iteration step, between these six classes. 

\begin{figure}
\centering
\fbox{\includegraphics[width=0.46\textwidth]{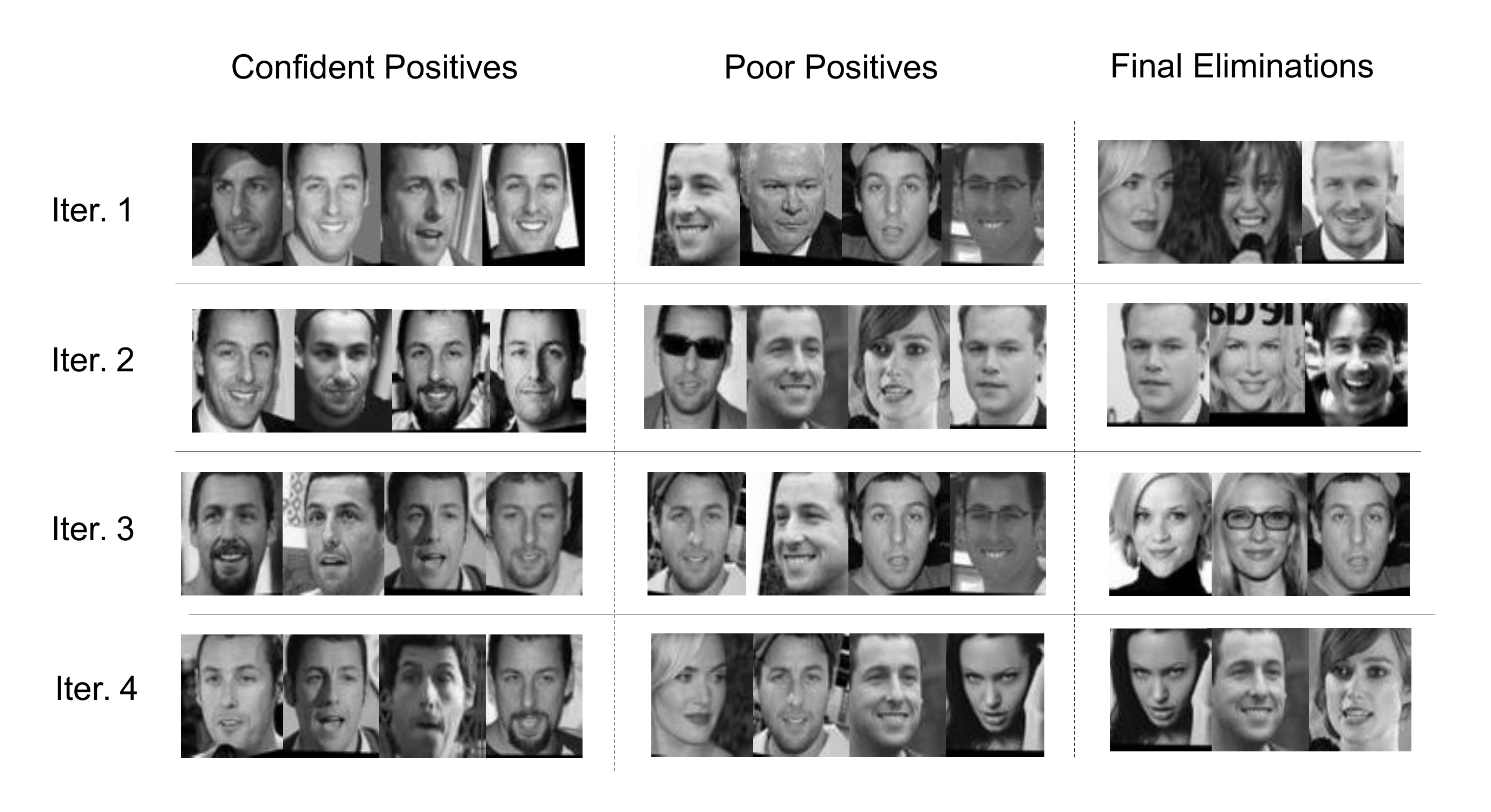}}
\fbox{\includegraphics[width=0.46\textwidth]{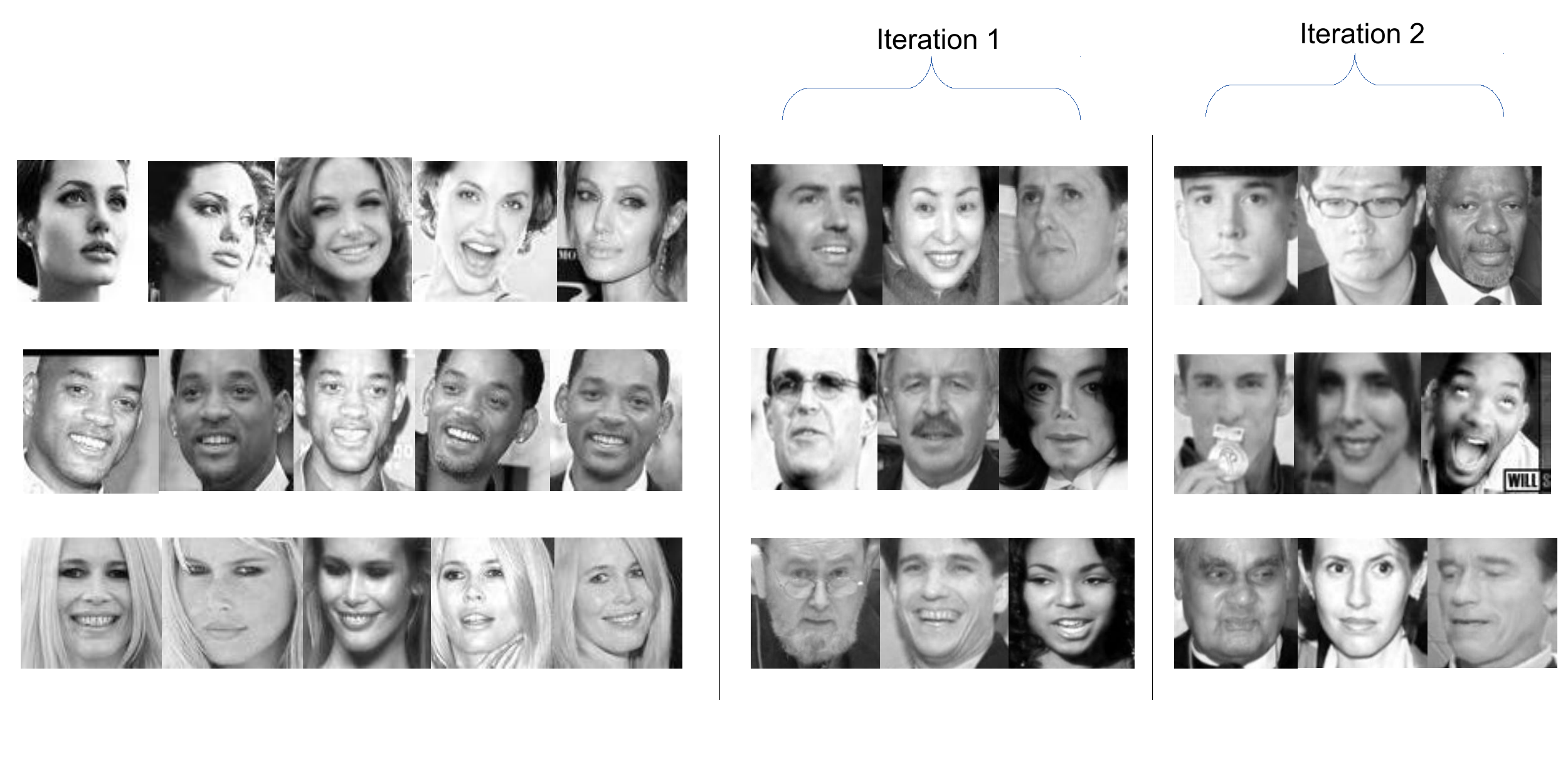}}
\caption{(Top:) Some of the instances selected for $C^+$, $C^-$ and $O$ for iterations $t=1\ldots 4$. 
(Bottom:) Samples for final model faces and outliers found in the first two iterations.}
\label{fig:iterations}
\end{figure}

\begin{figure}
\begin{center}
\begin{minipage}[b]{0.32\textwidth}
\centering
\includegraphics[width=\textwidth]{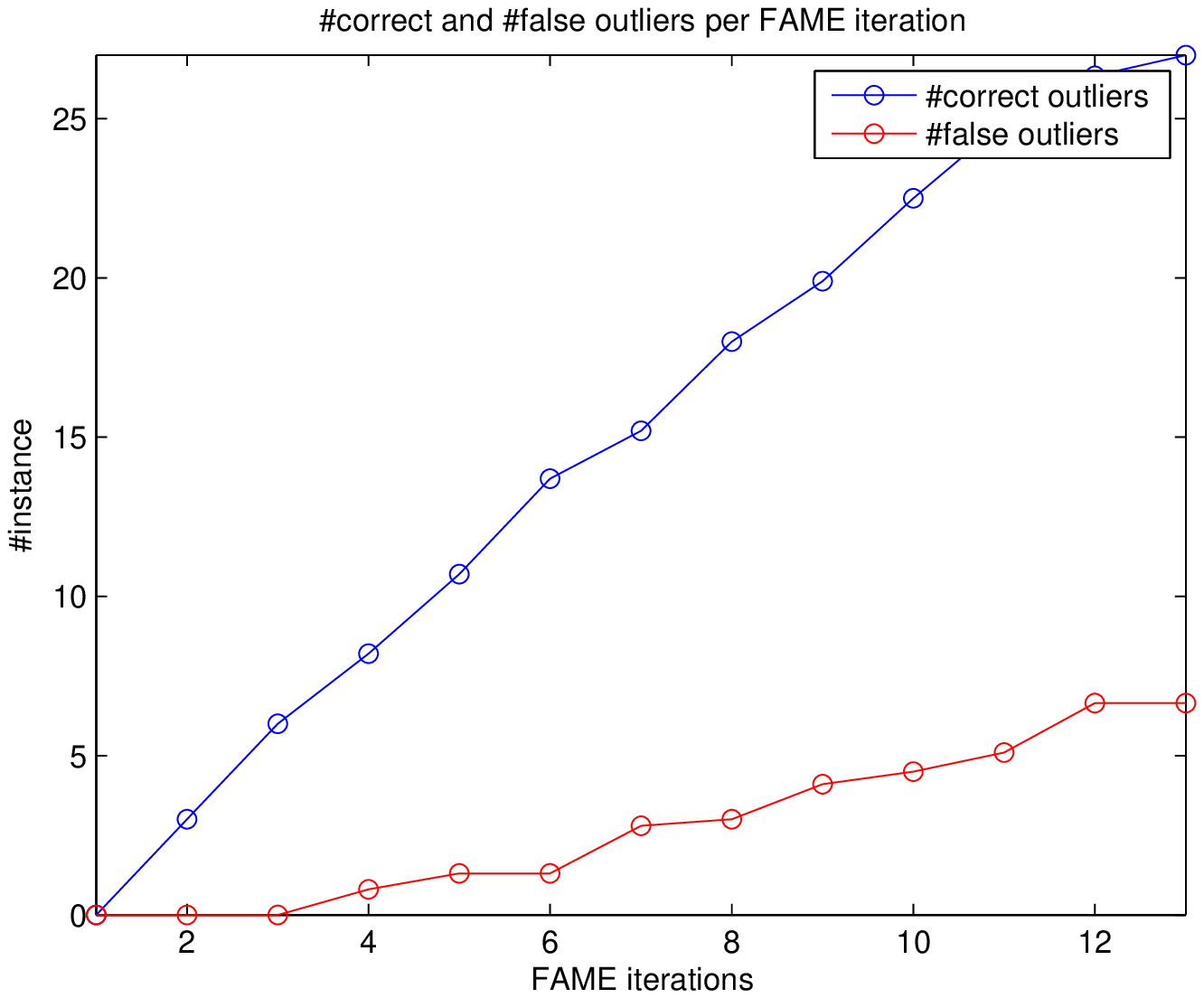}
\end{minipage}
\end{center}
\begin{center}
\begin{minipage}[b]{0.32\textwidth}
\centering
\includegraphics[width=\textwidth]{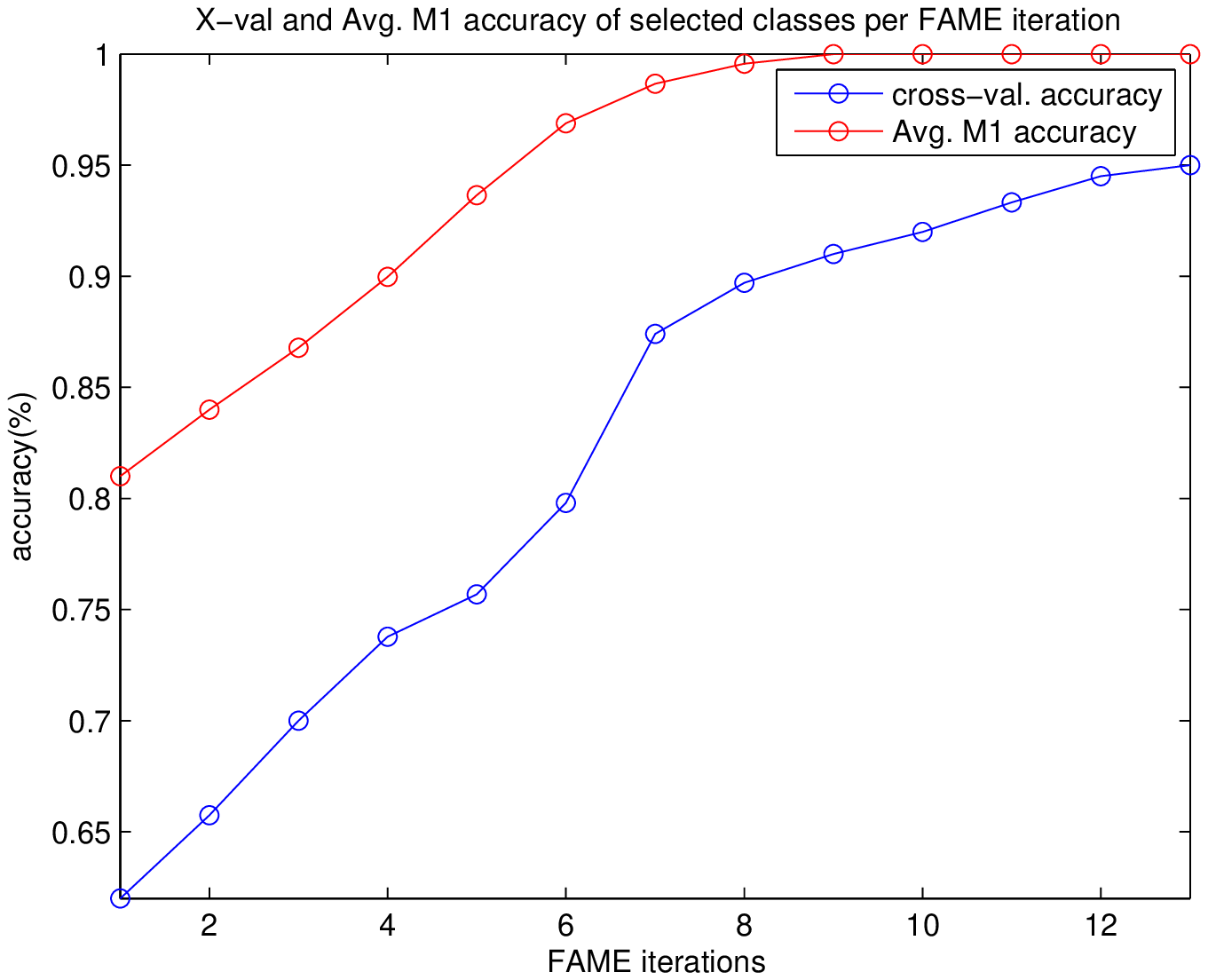}
\end{minipage}
\end{center}
\begin{center}
\begin{minipage}[b]{0.32\textwidth}
\centering
\includegraphics[width=\textwidth]{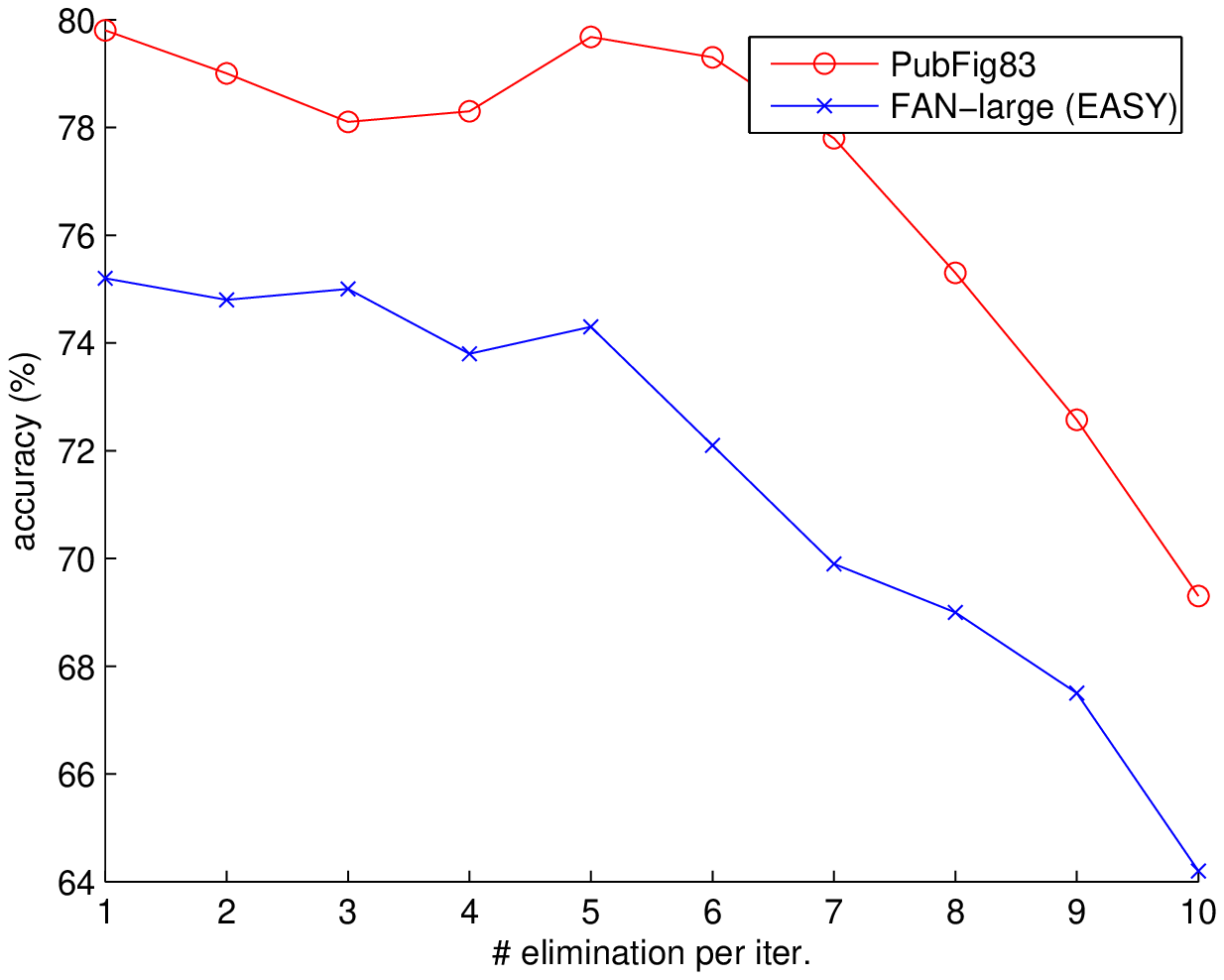}
\end{minipage}
\end{center}
\label{fig:controlled_plots}
\caption{ Evaluations on PubFig83 dataset.
(a) Total number of correct versus false outlier detections until FAME finds all the outliers for all the classes. We stop FAME for the saturated classes before the end of the plot. 
(b) Cross-validation and M1 accuracies as the algorithm proceeds. This shows the salient correlation between cross-validation classifier and the M1 models, without M1 models incurring over-fitting.
(c) Effect of number of outliers removed at each iteration.}
\end{figure}

Figure\ref{fig:iterations} helps to visualise the model evolution in FAME. As shown on the left, at each iteration dataset is divided into candidate positives and possible negatives: candidate positives are selected as the most representative instances of the class and true outliers are found among the possible negatives. As shown on the right, FAME is able to learn models from noisy weakly label sets, while eliminating the outliers at successive iterations for a variety of people.
As Figure\ref{fig:controlled_plots}-(a) shows with the increasing number of iterations, more outliers are eliminated. Although some correct instances are also eliminated, the ratio is very low compared to the spurious instances. Moreover, our observations show that the eliminated positive examples are usually not in good quality and therefore their elimination from the final model is not harmful but rather helpful as supported with the results in Figure\ref{fig:controlled_plots}-(b). 
As seen in Figure\ref{fig:controlled_plots}-(c) , we can achieve up to 75.2 on FAN-Large (EASY) and  79.8 on 
PubFig83 by removing one outlier at each iteration: we prefer to eliminate five outliers for the efficiency.
 
\begin{table}
\caption{ (Left:) This table compares the performances obtained with different features on PubFig83 dataset with the models learned from web. As the figure suggests, even LBP filters are not competitive with raw-pixel filters,  its textural information is subsidiary to raw-pixel filters with increasing performance. (Right:) Accuracy versus number of centroids $k$.  }
\vspace{0.3cm}
\begin{minipage}[b]{0.12\textwidth}
\begin{center}
\scalebox{0.60}{
	\begin{tabular}{|l|c|}
          \hline
           Feature & Accuracy \\ \hline
           LBP filters & 60.7 \\ \hline
           raw-pixel filters & 71.6 \\ \hline
           LBP+raw-pixel filters & 79.3 \\ \hline
   \end{tabular}}   
\end{center}
\end{minipage}
\hspace{1.5cm}
\begin{minipage}[b]{0.12\textwidth}
\begin{center}
\scalebox{0.65}{
  \begin{tabular}{|l|c|c|c|c|}
          \hline
          Num. Centroids & 1500 & 2000 & 2400\\ \hline
          Accuracy(\%) & 84.9 & 88.60 & 90.75\\ \hline
%          \cite{ozcan2011large} & 58.3 & 31.9 & 55.1 & 51.8 & - \\ \hline
  \end{tabular}}
  \end{center}
\end{minipage}
\label{fig:outliers}
\end{table}

We compare FAME with baseline method 
that learns models from the raw collection gathered through querying the name without any pruning. As seen in Table\ref{fan_large_results}  with one vs all L1 norm Linear SVM model on the raw data, the performance is very low on all datasets. Note that, on the datasets FAN-Large EASY and ALL, as well as PubFig83, we learn the models from web images and tested them on these novel datasets for the same categories. We also divided the collected Bing images into two subsets to test the effect of training and testing on the same type of dataset. FAME leads encouraging results even the model is susceptible to domain shifting problem, with a significant improvement over  baseline. 

The most similar data handling approach to ours is the method of Singh \etal \cite{singh2012unsupervised}, although there are important differences.  First, \cite{singh2012unsupervised} clusters the data to capture intra-cluster variance and uncover the representative instances. However, it requires to decide the optimal cluster number in advance  and divides the problem into multiple homologous pieces which need to be solved separately. This increase the complexity of the proposed system. 

Second difference lies in the philosophy. They aim to discover representative and discriminative set of instances whereas we aim to prune spurious ones. Hence, they need to keep all vast negative instances on memory but we can  sample different subsets of global negatives and find corresponding outlier instances. It provides faster and easier way of data pruning. They divide each class into two sets and apply their scheme by interchanging data after each iteration like in the case of co-training learning procedure.  Nevertheless,  co-training  demands large number of instances for reliable results. In our methodology, we prefer to use all the class data at once in our particular scheme. We evaluate the method of Sing \etal on the same datasets, and show that FAME is superior to their method (see Table\ref{fan_large_results}). We use the released code by  Singh \etal \cite{singh2012unsupervised} with up-limit settings that our resources allow.

To test the effectiveness of the proposed linear regression based model learning, we also compare our results by using only the $M^1$ model (FAME-M1) and using SVM for classification (FAME-SVM). As shown in Table\ref{fan_large_results}, all FAME models outperforms the baseline method as well as the method of  \cite{singh2012unsupervised} with a large improvement with the proposed LR model.

\begin{table}
\caption{Accuracies (\%) on FAN-Large \cite{ozcan2011large}  (EASY and ALL), PubFig83  and on the held-out set of our Bing data collection. There are three alternative FAME implementations. FAME-M1 uses only the model M1 which removes  instances regarding global negatives. FAME-SVM uses SVM in training and FAME-LR is the proposed method using linear regression.}
\begin{center}
	\scalebox{0.7}{
  \begin{tabular}{|l|c|c|c|c|}
          \hline
           - & Bing & FAN-Large (EASY) & FAN-Large (ALL) & PubFig83 \\ \hline
          Baseline & 62.5 & 56.5 & 52.7 &  52.8 \\ \hline
          Singh \etal \cite{singh2012unsupervised} & 74.7 & 65.9 & 62.3 &  71.4 \\ \hline
          FAME-M1 & 78.6 & 68.3 & 60.2 & 71.7 \\ \hline
          FAME-SVM & 81.4 & 73.1 & 65.4 & 76.8 \\ \hline
          FAME-LR & 83.7 & 74.3 & 67.1 & 79.3 \\ \hline
%          \cite{ozcan2011large} & 58.3 & 31.9 & 55.1 & 51.8 & - \\ \hline
  \end{tabular}
  }
\end{center}
\label{fan_large_results}
\end{table}

Finally, we compare the performance of FAME on the benchmark PubFig83 dataset with the other state-of-the-art studies on face identification. In this case, unlike the previous experiments where we learned the models from noisy images, in order to make a fair comparison we learned the models from the same dataset. As seen in Table\ref{table:PubFig83} FAME achieves the best accuracy in this setting. Referring back to  Table\ref{fan_large_results} even with the domain adaptation setting where the model is learned from the noisy web images our results are comparable to the most recent studies on face identification  that train and test on the same dataset. Note that, the method of Pinto \etal \cite{Pinto-CVPR2011} is similar to our classification pipeline but we prefer to learn the filters in an unsupervised way with the method of Coastes \etal \cite{coates2011analysis}..

\begin{table}
\caption{Accuracies (\%) of face identification methods on PubFig83. \cite{pinto2011scaling} proposes single layer (S) and multi-layer (M) architectures. {\tt face.com} API is also experienced in \cite{pinto2011scaling}.  Note that, here FAME is learned from the same dataset. }
\begin{center}
	\scalebox{0.55}{
	  \begin{tabular}{|l|c|c|c|c|c|}
	  		\hline
	  		 Method & Pinto \etal \cite{pinto2011scaling} (S) & Pinto et al.\cite{pinto2011scaling}(M) & face.com \cite{pinto2011scaling} & Becker \etal \cite{becker2013evaluating} &  $FAME$ \\ \hline
	  		 Accuracy & 75.6 & 87.1 & 82.1 & 85.9 & 90.75   \\ \hline
	%  		\cite{ozcan2011large} & 58.3 & 31.9 & 55.1 & 51.8 & - \\ \hline
	  \end{tabular}
  }
\end{center}
\label{table:PubFig83}
\end{table}

%\section{Discussion on the relevant methods}
%In \cite{Li-IJCV2010} OPTIMOL framework introduces similar approach to FAME. They suggest to use a manually selected seed images or top returned images by a state-of-art search engine. OPTIMOL extends the dataset with new images by the guidance of seed images. However, such an approach might overlook intra-class variance so causes to degrade generalization ability of the system. Even worse, seed set might cause concept shift owing to impostor members. OPTIMOL does not provide any recovery method for underlined problems. On the other side, FAME finds seed images for eact iteration without any extraneous help by using the discriminativeness score against global negative images.Therefore it is more robust and automated in relation to OPTIMOL

\section{Summary and future work}
We propose a novel method to prune the web images collected for a query to learn models to be used for classification on novel datasets. We rely on large number of negative instances
in selecting a set of good instances which are then used to learn models to eliminate the bad ones. The proposed method outperforms the baseline and is comparable to state-of-the-art methods even within the difficulties of domain adaptation.
Although the proposed method is tested for identification of faces, it is a general method that could be used for other domains as we aim to attack as our future work.

%Our proposed schema relies onto very large set of negative instances that are gathered randomly. In that way, we are able to learn some of discerning features of our class against the world, with the assumption that on the whole, collected class instances composed by the class related  members so that we can detect irrelevant or outlier instances by believing the majority. 

\bibliographystyle{plain}
\bibliography{fame_arxiv_2014}
\end{document}